\documentclass[11pt]{article}
\usepackage{eamt24}
\usepackage{times}
\usepackage{url}
\usepackage{latexsym}
\usepackage{xcolor}
\usepackage[most]{tcolorbox}
\usepackage{booktabs}
\usepackage{csquotes}
\usepackage{microtype}
\usepackage{inconsolata}
\usepackage{multirow}
\usepackage[algoruled]{algorithm2e}
\SetKwComment{Comment}{\# }{}

\usepackage[small,bf]{caption} 
\usepackage{todonotes}

\title{Prompting Large Language Models with Human Error Markings for Self-Correcting Machine Translation}

\author{Nathaniel Berger$^{a}$\thanks{\hspace{1ex}The work was done as part of an SAP sponsored PhD project of the first author.}~~and Stefan Riezler$^{ab}$ \\
  Computational Linguistics$^a$ \& IWR$^{b}$ \\
  Heidelberg University \\
  69120 Heidelberg, Germany \\
  {\tt berger@cl.uni-heidelberg.de}  \\
  {\tt riezler@cl.uni-heidelberg.de} \And
  Miriam Exel$^c$ and Matthias Huck$^c$ \\
  SAP SE$^c$ \\
  Dietmar-Hopp-Allee 16 \\
  69190 Walldorf, Germany \\
  {\tt miriam.exel@sap.com}\\
  {\tt matthias.huck@sap.com}}

\date{}

\begin{document}
\maketitle
\begin{abstract}
While large language models (LLMs) pre-trained on massive amounts of unpaired language data have reached the state-of-the-art in machine translation (MT) of general domain texts, post-editing (PE) is still required to correct errors and to enhance term translation quality in specialized domains.  In this paper we present a pilot study of enhancing translation memories (TM) produced by PE (source segments, machine translations, and reference translations, henceforth called PE-TM) for the needs of correct and consistent term translation in technical domains. 

We investigate a light-weight two-step scenario where, at inference time, a human translator marks errors in the first translation step,  and in a second step a few similar examples are extracted from the PE-TM to prompt an LLM. Our experiment shows that the additional effort of augmenting translations with human error markings guides the LLM to focus on a correction of the marked errors, yielding consistent improvements over automatic PE (APE) and MT from scratch.
\end{abstract}

\begin{figure*}[t!]
\begin{tcolorbox}[minipage,colback=lightgray,arc=0pt,outer arc=0pt]
Read the English text and the German translation hypothesis and then correct the output. \\
Incorrect words are inside of tags '\textbf{$<$bad$>$ $<$/bad$>$}'. Please use this feedback in your correction. \\
If the hypothesis is already correct, do not make any changes.\\

\textcolor{teal}{English: This environment variable can also be used to make sure that other operations are working on uploaded files, as well.}\\
\textcolor{blue}{Hypothesis: Dieses \textbf{$<$bad$>$} Umweltvariable \textbf{$<$/bad$>$} kann auch verwendet werden , um sicherzustellen , dass andere \textbf{$<$bad$>$} Operationen auf \textbf{$<$/bad$>$} hochgeladene Dateien \textbf{$<$bad$>$} arbeiten \textbf{$<$/bad$>$} .}\\
\textcolor{purple}{German: Mittels dieser Umgebungsvariable kann auch sichergestellt werden, dass auch andere Operationen an hochgeladenen Dateien arbeiten können.}\\

~\\
\textcolor{teal}{English: Some important environment variables used by KDE}\\
\textcolor{blue}{Hypothesis: Einige wichtige \textbf{$<$bad$>$} Umweltvariablen \textbf{$<$/bad$>$} , die von KDE verwendet werden}\\
\textcolor{purple}{\textbf{German: Einige wichtige Umgebungsvariablen, die von KDE verwendet werden}}\\
\end{tcolorbox}
\caption{Example of a 1-shot prompt for English-to-German Translation. Error markings are inside bold faced tags \textbf{$<$bad$>$ $<$/bad$>$}. The demonstration example consists of a source segment in English (in green), a translation hypothesis in German (in blue), and a correction (in red). The test example shows a correction of the translation of "environment variable" from "Umweltvariable" into "Umgebungsvariable" learned by the LLM (in bold-faced red).}
\label{fig:prompt}
\end{figure*}

\section{Introduction}

Technical translation at large enterprises involves a large number of translation domains, for which translation memories and terminologies need to be maintained to support multi-domain MT systems and human post-editors in producing contextually adequate and consistent translation of technical terms \cite{ExelETAL:20}. In this paper, we ask if ongoing human post-editing efforts that produce large databases consisting of source segments, machine translations, and reference translations, can be enhanced by light-weight human error markings. 
This could then be used to teach a translation system a focused self-correction of marked erroneous tokens from similar examples with error markings and corrections found in the PE-TM. Such a setup could complement translation memories and terminology databases by up-to-date and domain-specific information in the PE-TM, and be used in a scenario where a user marks errors in MT hypotheses. In-context examples with high source-side similarity are then extracted from the PE-TM to prompt an LLM to focus on a correction of the marked error interactively. 

We present a pilot study where we construct a PE-TM for the IT domain, which is augmented by human error markings on machine translations. While for training purposes, error markings for the PE-TM could be obtained by automatic matching against human post-edits, this cannot be done at test time. We envisage a scenario where the error markings in the PE-TM are obtained by direct human annotation, simulating a realistic setup where a user only performs the light-weight task of error marking at test time. 
Such a scenario could be feedback collection in the publishing of raw-MT. Raw-MT could be shown to end-users who, if they notice an error in the translation, proceed to annotate tokens in the translation they perceive to be incorrect. The translation would then be flagged for review by a human translator, who then post-edits the translation and publishes their correction. This process results in the creation of (source, hypothesis, post-edit) triples with annotations for the PE-TM. 
This PE-TM is used to provide in-context examples for LLM correction of annotated translation hypotheses. For example, the end-user who annotates raw-MT could then immediately be shown a new translation that takes the error markings into account. 

The results of our study show that selecting in-context examples based on similarity of source-side embeddings and providing error markings on hypotheses lets the LLM infer focused corrections of marked errors. Furthermore, overall translation quality is improved over few-shot prompt-based translation and over automatic post-editing. An example 1-shot prompt and error-marked output is given in Figure \ref{fig:prompt}.

\section{Prior Work}

The last year has seen a progression of the translation capabilities of decoder-only LLMs, pre-trained on unpaired language data, from lagging behind supervised systems \cite{VilarETAL:23} to matching their translation quality \cite{GarciaETAL:23}, with only 5 examples of high-quality translation data used for in-context learning. However, MT in specialized domains still requires translation post-editing in order to correct errors and to enhance term translation quality. Raunak et al., \shortcite{RaunakETAL:23} recently showed that very large LLMs \cite{OpenAI:23} can perform zero-shot automatic translation post-editing for general domain data, at the price of hallucinated edits. This makes this setup impractical if high precision in domain-specific translation is key. For these purposes, manually crafted glossaries \cite{VidalETAL:22}, dictionaries extracted in a separate step of unsupervised word-alignment \cite{GhazvininejadETAL:23}, or translation memories accessed with fuzzy matching \cite{MoslemETAL:23,HoangETAL:23}, have been used to aid prompt-based MT. Our approach combines PE-TMs with light-weight human error markings, achieving improvements over both APE and MT from scratch.

The standard paradigm to incorporate token-level human error markings as learning signal is an adaptation of supervised learning from post-edits (see, for example, \newcite{TurchiETAL:17}) by penalizing erroneous tokens and rewarding correct tokens in a weighted maximum-likelihood objective 
\cite{MarieMax:15,DomingoETAL:17,PetrushkovETAL:18,LamETAL:19,KreutzerETAL:20,BergerETAL:23}. Most approaches are conceptualized as fine-tuning applications, with error markings obtained by automatic matching against human post-edits or by direct human annotation. The approach that is closest to our work is QuickEdit \cite{GrangierAuli:18}. They train a model with separate encoders for source and error-marked hypothesis in order to improve upon the initial hypothesis by avoiding the marked tokens. Similar to our approach, QuickEdit requires error-markings at inference time.
While QuickEdit relies on supervised learning, our approach succeeds in teaching an LLM to avoid marked tokens from a few demonstration examples of similar error patterns. 

More recent work by Xu et al., \shortcite{XuETAL:23pinpoint} successfully uses feedback in form of error type and location that is predicted by a learned error pinpoint model. Their work focuses on general domain translation and quality-estimation type feedback, in difference to the focused error markings on technical terms that we are interested in. We plan an extension of our work in the direction of using learned error markings in future work.

Our work is furthermore related to the more general issue of self-correction capabilities of LLMs. Similar to the findings of Huang et al., \shortcite{HuangETAL:23selfcorrect}, our work shows that in order to qualify as a correction rather than a mere change, automatic self-correction in LLMs \cite{MadaanETAL:23,PanETAL:23} needs to be guided by an oracle. 
In our case, the oracle consists of feedback on the errors in translation outputs of an LLM, combined with a few examples of similar errors and their reference translations.

\section{Data and Models}

We collected English and German parallel data from open source software documentation and localization available on OPUS \cite{Tiedemann:12}, as this data comes closest to our domain of interest. We concatenated data from \textit{GNOME}, \textit{KDE4}, \textit{KDEdoc}, \textit{PHP}, and \textit{Ubuntu} to create our data set and filtered them with the following methods: we removed those segments containing fewer than five words or more than 25, those identified by fastText \cite{JoulinETAL:17} as the wrong language, those with more than 20\% of characters being non-alphanumeric, and those containing personally identifiable information. Of the remaining data, we selected a subset of $1,500$ examples.

For the purposes of this experiment, we were interested in models that support prompt-based interaction. Furthermore, we are interested in the scenario where users use their judgment to guide a model towards a better translation based upon its original translation. Large language models lend themselves well to this interaction because the same model can be used with prompt-based interaction to produce the original translations as well as for providing extra information to aid in correction. These considerations decide in favor of using an LLM over a traditional encoder-decoder based model typically used in production scenarios. Therefore, we examine if the model that produced the hypothesis, Llama 13B \cite{TouvronETAL:23}, can leverage feedback to correct its own mistakes. In addition to choosing this model because it supports prompting, it runs on a single GPU\footnote{For all Llama 13B experiments, we use a single Nvidia A40 GPU with 48GB VRAM on a shared server}, and the model will remain available in the future for reproducibility. The Llama model was converted to Huggingface Transformers \cite{WolfETAL:20} format for inference. We found that Llama 13B frequently copies hypotheses including the error tags to its output as it was instructed not to make changes if the hypothesis is acceptable as-is. We therefore post-processed outputs by removing tags by regex. 

Additionally, we test GPT-3.5\footnote{GPT-3.5-Turbo-0613 was used for all experiments involving OpenAI's GPT models in this paper} in order to test a model larger than we can locally run. We use the "ChatCompletion" API, send the query as the 'user' message, and set the temperature to $0$. All hypotheses were generated with the above models using greedy decoding. 

\begin{figure*}[t!]
\begin{tcolorbox}[minipage,colback=lightgray,arc=0pt,outer arc=0pt]
\small
You will receive an annotation task called “Error Annotation”; the goal of this task is to mark each word in the machine translation as correct or incorrect. By default, all words are considered correct. By clicking on target words, they are marked as incorrect.\\

Here are the instructions in more detail:\\
You will be shown an English source sentence above and its machine translation into German below.
\begin{itemize}
    \item	Begin by reading the source sentence and then reading the translation.
    \item	Consider which words would need to be deleted or changed in order to arrive at a correct translation.
    \item	Mark the incorrect words of the translation by clicking on them.
    \item	Clicking on the word causes a blue border to appear around the word. This word is now marked as incorrect.
    \item	Clicking a second time will remove the blue border and it is now marked correct.
    \item	Once all the incorrect words have a blue border, click on the “Next” button near the top of the page.
    \item	Markings should be kept minimal. Mark only those terms that you would edit or delete in a post-editing scenario.
    \item	If you would have to move a word to a different location, such as shifting a verb to the end of the sentence, mark it as incorrect.
\end{itemize}

If the translation contains no correct words or the source words are translated word by word but do not make sense together, mark them all as incorrect.\\

If the translation is correct as-is, proceed to the next annotation item.\\

If you cannot judge the quality of the translation because the source sentence is not comprehensible, or you are lacking domain knowledge to annotate wrong words, click the Skip button (to the right of the “Next” button) and then proceed to the next sentence.\\

The source sentences are taken from open-source software projects and documentation while the translations are produced by a generic machine translation system.
\end{tcolorbox}
\caption{Instructions given to annotators on how to mark errors in sentences, including how to use the interface and desired marking behavior}
\label{fig:annotator-instructions}
\end{figure*}

\section{Feedback Collection}

In order to simulate the previously proposed scenario of an end-user who annotates raw-MT errors, we turn to paid annotators. We generate translations of English source sentences and provide them with only the source and the hypothesis, as would be the case when getting feedback on raw-MT. These are then paired with references to create our PE-TM.

\subsection{Human Annotation}

We hired three professional translators with expertise in the IT domain as annotators to provide token-level feedback on the translation hypotheses. Token-level feedback consisted of per-token binary quality judgements, OK/BAD. Annotators were provided English source sentences and German hypothesis translations in a custom annotation interface. Each token in the hypothesis was a button in the annotation interface and annotators were instructed to click on incorrect tokens to mark errors. Unmarked tokens were assumed OK. Additionally, they were instructed to keep markings minimal and only mark tokens that would be edited or deleted during post-editing. 

Complete instructions are shown in Figure \ref{fig:annotator-instructions}. Annotators could skip examples but must provide a reason. Reasons for skipping examples were "Source Incomprehensible", "Source Ambiguous", "Missing Knowledge", and "Other". 

Annotation was split into two phases. Phase one was a trial run where all three annotators annotated the same 50 examples. In phase two, each annotator was given their own non-overlapping block of 500 source and hypothesis pairs. The phase one examples were used to compute summary statistics of annotation behavior, agreement coefficients, and to calibrate our instructions.

After phase two, filtering out skipped examples or those without any BAD markings yields a data set of 982 examples. We split this data set into two subsets; one set of size 492 for in-context examples and a set of 490 for test examples.

\subsection{Annotation Statistics}

Annotator 1 selected "Source Ambiguous" as the reason for skipping once and "Missing Knowledge" the other six times. Annotator 2 selected "Source Incomprehensible" for their skip.
After removing the items skipped by any annotator, we have 43 examples that were annotated by all three. 

Using the remaining common examples from phase one, we calculate the percentage of tokens marked per sentence and use that as a sentence-level quality judgment. This is then used to calculate Krippendorff's alpha \cite{KrippendorffReliability:04} to determine if our annotators agree on overall translation quality. We also calculate alpha on the token level OK/BAD annotations.

\begin{table}[t!]
\centering
\begin{tabular}{rrrr}
\toprule
Annotator & 1 & 2 & 3 \\
\midrule
Percent Marked on Average & 0.25 & 0.17 & 0.17\\
SD of Percent Marked & 0.28 & 0.18 & 0.19 \\
\bottomrule
\end{tabular}
\caption{Marking behaviors of each annotator in terms of percent of tokens marked in the trial annotation.}
\label{tab:percent-marked-trial}
\end{table}

\begin{table}[t!]
\centering
\begin{tabular}{rrr}
\toprule
Annotator & 2 & 3 \\
\midrule
1 & 0.258 & 0.481 \\
2 & $\emptyset$ & 0.222 \\
\bottomrule
\end{tabular}
\caption{Inter-annotator agreement for percentage marked per sentence, given by Krippendorff's Alpha.}
\label{tab:agreementpercentagemarked}
\end{table}

\begin{table}[t!]
\centering
\begin{tabular}{rrr}
\toprule
Annotator & 2 & 3 \\
\midrule
1 & 0.445 & 0.531 \\
2 & $\emptyset$ & 0.433 \\
\bottomrule
\end{tabular}
\caption{Inter-annotator agreement for token classification, given by Krippendorff's Alpha.}
\label{tab:agreement-token-classification}
\end{table}

We calculated pair-wise Krippendorff's alpha in addition to the average agreement for both the sentence-level percentage marked and token-level annotations. The average amount of tokens marked for the unskipped sentences is visible in Table \ref{tab:percent-marked-trial}. Pairwise Krippendorff's alphas for percentage marked is visible in Table~\ref{tab:agreementpercentagemarked}, while pairwise agreement for token classification is in Table~\ref{tab:agreement-token-classification}. Average agreement for the percentage marked is $\alpha = 0.306$ and for token classification $\alpha = 0.466$. This suggests that, while agreement about overall sentence quality is not high, the reliability of classifying each token in the hypothesis is higher. These results were used to calibrate with the annotators after looking over the annotations made by each individual.

After calibration, we then assigned each annotator their block of 500 examples to annotate. Annotator 1 skipped 6 of the 500 sentences and annotator 2 skipped 20. Percentage marked was lower for annotators 1 and 3 during the full annotation as more sentences were left completely unmarked. Annotator 1 left 36\% of sentences unmarked; annotator 2 left 23\%; and annotator 3 left 38\%. The percentage that was marked per sentence was also reduced after calibration, as shown in Table \ref{tab:percent-marked-full}.

\begin{table}[t!]
\centering
\begin{tabular}{rrrr}
\toprule
Annotator & 1 & 2 & 3 \\
\midrule
Percent Marked on Average & 0.10 & 0.19 & 0.09\\
SD of Percent Marked & 0.10 & 0.15 & 0.1 \\
\hline
\end{tabular}
\caption{Marking behaviors of each annotator in terms of percent of tokens marked in the final annotation.}
\label{tab:percent-marked-full}
\end{table}

\section{Experiments}

\subsection{Experimental Setup}

\begin{table*}[t!]
\centering
\begin{tabular}{r|ccccc}
\toprule
    Condition & BLEU & TER & ME & UE & \% Correct ME \\
   \midrule
 Original Hyps & 28.92 & 55.12 & N.A. & N.A. & N.A. \\
  \midrule
 MT (Llama/GPT) & 29.83/38.61 & 55.97/49.21 & N.A. &  N.A. & N.A. \\
 APE (Llama/GPT) & 29.79/39.09 & 54.56/48.37 & 7.30/76.70 & 1.76/15.85 & 32\% / N.A. \\
 MRK (Llama/GPT) & 30.09/39.31 & 54.70/48.32 & 14.76/78.36 & 3.60/13.90 & 67\% / N.A. \\
\bottomrule
\end{tabular}
\caption{Results for both Llama 13B and GPT 3.5 across all metrics and translation scenarios (ME = Marking Edits, UE = Unmarking Edits, \% Correct ME = Percentage of correct ME in manual evaluation).}
\label{tab:metric-results}
\end{table*}

Using the annotated data, we considered three machine translation tasks: Machine translation from scratch (\textit{MT}); Automatic Post-editing (\textit{APE}); and Post-Editing with error markings (\textit{MRK}). Instructions were written for the LLM for each task and, for each example in the inference set, five examples were retrieved from the in-context example pool. We retrieve the most similar examples by using cosine similarity over SentenceTransformers \cite{ReimersETAL:19} embeddings computed on source sentences only\footnote{We used the model \textit{all-MiniLM-L6-v2} and retrieved the examples with the highest cosine similarity.}. 

In \textit{MT}, models were prompted to 

\begin{displayquote}
    Translate English to German.
\end{displayquote}

and were shown five (source, reference) pairs. Full prompts can be found in the appendix \ref{sec:example-prompts}. For the \textit{APE} task, models were prompted to

\begin{displayquote}
    Read the English text and the German translation hypothesis and then correct the output. If the hypothesis is already correct, do not make any changes.
\end{displayquote} 

With this prompt, the models were given triples of (source, hypothesis, reference) with the hypothesis from our annotated data set and the reference coming from the parallel data.

In the \textit{MRK} scenario, models were prompted to

\begin{displayquote}
    Read the English text and the German translation hypothesis and then correct the output. Incorrect words are inside of tags '$<$bad$>$ $<$/bad$>$'. Please use this feedback in your correction.  If the hypothesis is already correct, do not make any changes.
\end{displayquote}

As with the \textit{APE} prompt, models were given (source, hypothesis, reference) triples with the tokens that were marked as bad during annotation inside of XML-style tags, \texttt{$<$bad$>$$<$/bad$>$}. We decided that giving the error markings as in-line tags would be easier for the model to parse and integrate in its output than including another line where errors would be indicated further away from the corresponding tokens.

\subsection{Metrics}

We evaluate the models' new hypotheses with a suite of metrics to check for token level matches, semantic similarity, and error marking usage. We use the token based metrics BLEU\footnote{BLEU signature: \texttt{nrefs:1 | case:mixed | eff:no | tok:13a | smooth:exp | version:2.4.0}} \cite{Papineni:02} and TER\footnote{TER Signature \texttt{nrefs:1 | case:lc | tok:tercom | norm:no | punct:yes | asian:no | version:2.4.0}} \cite{SnoverETAL:06} as implemented in SacreBLEU \cite{Post:18}. We did not include the popular neural metric COMET \cite{ReiETAL:20} since is not sensitive to the individual token changes \cite{GlushkovaETAL:23} that we ask the LLMs to perform. 

In addition to these metrics, we also implement our own to see how well the models are at recognizing and making edits to errors. These are called \textit{marking edit} (ME) and \textit{unmarking edit} (UE). We perform a word-level diff in order to see which words need to be edited or deleted in the original hypothesis to arrive at the new hypothesis. Combining this with the error markings allows us to examine if edits were made to the tokens error-marked by the annotators (ME), or if tokens that were otherwise OK were changed (UE). 

The ME and UE metrics, however, cannot tell if the edits were correct, only that they were made. To determine if the edits correctly fix the errors, we performed a manual evaluation on the marking edits produced by Llama 13B in both the \textit{APE} and \textit{MRK} settings. 
Three of the authors contributed to the evaluation. Two are native speakers of German and fluent in English while one is a native speaker of English and fluent in German. We selected 100 sentences with the most marking edits. Edits were evaluated in terms of their correctness, with a subjective yes or no answer given for the entire sentence.

\section{Results}

We show results across metrics for Llama 13B and GPT-3.5 in Table \ref{tab:metric-results}. Including error markings as input increases the frequency with which the models edits the marked tokens. For Llama 13B, we see editing rates for marked tokens double from $7.30$ to $14.76$. This suggests that, even after being asked to correct the hypotheses, Llama 13B finds its own outputs as acceptable translations. When errors are specifically pointed out to the model, it is much more capable of self-correcting errors. 

Llama 13B nominally improves BLEU scores over the original hypotheses score ($28.92$) in all scenarios with \textit{MRK} in the lead with $30.09$, \textit{MT} in second with $29.83$\footnote{\textit{MT} is able to surpass the original hypotheses with Llama 13B because the annotated hypotheses were generated with the same 5 examples for all inference segments while \textit{MT} retrieved similar examples for each test segment.} and \textit{APE} with $29.79$. Nominal improvements over the original hypotheses are also found according to the TER metric, albeit only for \textit{APE} and \textit{MRK} scenarios.

The GPT model is already quite capable of finding errors in the hypotheses without error markings and the \textit{APE} outputs achieve marking edits of $76.70$ while \textit{MRK} has a slight improvement of $78.36$. Worth noting is the reduction in unmarking edits when prompting GPT with \textit{MRK}. \textit{MRK} reduces unmarking edits to $13.90$ from $15.85$ with \textit{APE}. This means that indicating specific errors can constrain the number of edits that the GPT model makes. Additionally, nominal improvements of BLEU and TER scores are found in the \textit{APE} and \textit{MRK} scenarios over \textit{MT} with GPT 3.5 as well. \textit{MRK} improves BLEU to $39.31$ from \textit{MT}'s $38.61$.

In the manual evaluation of marking edits, we found that \textit{APE} made correct edits 32\% of the time on average, while either making incorrect edits or not editing the rest. \textit{MRK} on the other hand was judged correct 67\% of the time on average. Agreement in terms of Krippendorff's Alpha for sentence level ratings of \textit{APE} is $\alpha = 0.82$, while for \textit{MRK} $\alpha = 0.55$. 
As \textit{APE} makes fewer edits overall, it is easier to classify as incorrect or not editing. For \textit{MRK} there was disagreement on how to handle partial edits or if not all markings were edited, requiring individual judgement by each evaluator.

\section{Conclusion}

We presented a pilot study to investigate the potential of augmenting a so-called PE-TM resource consisting of sources, machine translations, and human references, with human error markings in order to guide an LLM to self-correct marked erroneous term translations. We find that the LLM that produced the translation hypotheses identifies its own translations as correct, and therefore does not act on the instructions to correct errors. However, when prompted with error markings, the LLM learns to act on them, doubling the number of edits to marked tokens, with nearly 70\% of the edits being correct according to a human evaluation. 
In sum, our pilot study shows that the additional effort of error marking a machine translation at test time allows an LLM translation system to learn focused corrections on marked errors from similar examples extracted from a PE-TM, leading to 
improved translation quality over APE and MT. In future work, we will investigate learned models for error markings. These require larger TMs for reliable training of markings estimators, but also bear the promise of improved retrieval augmentation.

\section{Acknowledgements}

The second author acknowledges support by the state of Baden-Württemberg through bwHPC
and the German Research Foundation (DFG) through grant INST 35/1597-1 FUGG.

\bibliography{ref}
\bibliographystyle{eamt24}

\appendix
\section{Appendix}

\subsection{Example Prompts}
\label{sec:example-prompts}

An example of a prompt for \textit{MT}, \textit{APE}, and \textit{MRK} are in Figures \ref{fig:prompt-example-mt}, \ref{fig:prompt-example-ape}, and \ref{fig:prompt-example-mrk}, respectively.

\begin{figure*}[t!]
\begin{tcolorbox}[minipage,colback=lightgray,arc=0pt,outer arc=0pt]
Translate English to German.

\textcolor{teal}{English: Cookies are part of the HTTP header, so setcookie() must be called before any output is sent to the browser.}\\
\textcolor{purple}{German: Sie sind Bestandteil des HTTP-Headers, was bedeutet, dass die Funktion setcookie() aufgerufen werden muss, bevor irgendeine Ausgabe an den Browser erfolgt.}\\

\textcolor{teal}{English: session.use\_only\_cookies specifies whether the module will only use cookies to store the session id on the client side.}\\
\textcolor{purple}{German: session.use\_only\_cookies spezifiziert, ob das Modul nur Cookies verwendet, um die Session-ID clientseitig zu speichern.}\\

\textcolor{teal}{English: Note that SID is only defined if the client didn't send the right cookie.}\\
\textcolor{purple}{German: Beachten Sie, dass SID nur definiert ist, wenn vom Client nicht das richtige Cookie gesendet wurde.}\\

\textcolor{teal}{English: The server does not support the request type of the body.}\\
\textcolor{purple}{German: Der Server unterstützt den angeforderten Typ nicht.\%1: request type}\\

\textcolor{teal}{English: Must be in  active session on local console}\\
\textcolor{purple}{German: Nur in aktiver Sitzung auf lokaler Konsole}\\

\textcolor{teal}{English: Like other headers, cookies must be sent before any output from your script (this is a protocol restriction).}\\
\textcolor{purple}{German:}\\

\end{tcolorbox}

\caption{Example of 5-shot prompt for English-to-German Translation. Each demonstration example consists of a source segment in English (in green), and a reference translation (in red).}
\label{fig:prompt-example-mt}
\end{figure*}

\begin{figure*}[t!]
\begin{tcolorbox}[minipage,colback=lightgray,arc=0pt,outer arc=0pt]
Read the English text and the German translation hypothesis and then correct the output. If the hypothesis is already correct, do not make any changes. \\

\textcolor{teal}{English: Cookies are part of the HTTP header, so setcookie() must be called before any output is sent to the browser.}\\
\textcolor{blue}{Hypothesis: Cookies sind Teil des HTTP-Headers , deshalb muss setcookie() vor jedem Ausgabe-Output an den Browser aufgerufen werden .}\\
\textcolor{purple}{German: Sie sind Bestandteil des HTTP-Headers, was bedeutet, dass die Funktion setcookie() aufgerufen werden muss, bevor irgendeine Ausgabe an den Browser erfolgt.}\\

\textcolor{teal}{English: session.use\_only\_cookies specifies whether the module will only use cookies to store the session id on the client side.}\\
\textcolor{blue}{Hypothesis: session .use\_only\_cookies bestimmt , ob das Modul nur mit Cookies die Session-ID auf dem Client-Betriebssystem speichert .}\\
\textcolor{purple}{German: session.use\_only\_cookies spezifiziert, ob das Modul nur Cookies verwendet, um die Session-ID clientseitig zu speichern.}\\

\textcolor{teal}{English: Note that SID is only defined if the client didn't send the right cookie.}\\
\textcolor{blue}{Hypothesis: Beachtet , dass SID nur definiert ist , wenn der Client nicht den richtigen Cookie gesendet hat .}\\
\textcolor{purple}{German: Beachten Sie, dass SID nur definiert ist, wenn vom Client nicht das richtige Cookie gesendet wurde.}\\

\textcolor{teal}{English: The server does not support the request type of the body.}\\
\textcolor{blue}{Hypothesis: Der Server unterstützt nicht die Anforderungstyp der Body .}\\
\textcolor{purple}{German: Der Server unterstützt den angeforderten Typ nicht.\%1: request type}\\

\textcolor{teal}{English: Must be in \& active session on local console}\\
\textcolor{blue}{Hypothesis: Muss in \& aktiver Sitzung auf dem lokalen Konsole}\\
\textcolor{purple}{German: Nur in \& aktiver Sitzung auf lokaler Konsole}\\

\textcolor{teal}{English: Like other headers, cookies must be sent before any output from your script (this is a protocol restriction).}\\
\textcolor{blue}{Hypothesis: Wie andere Headern müssen Cookies vor jedem Ausgabe-Output ( dies ist eine Protokoll-Einschränkung ) gesendet werden .
}\\
\textcolor{purple}{German:}\\

\end{tcolorbox}

\caption{Example of 5-shot prompt for English-to-German Automatic Post-Editing (\textit{APE}). Each demonstration example consists of a source segment in English (in green), a translation hypothesis in German (in blue), and a reference translation (in red).}
\label{fig:prompt-example-ape}
\end{figure*}

\begin{figure*}[t!]
\begin{tcolorbox}[minipage,colback=lightgray,arc=0pt,outer arc=0pt]
Read the English text and the German translation hypothesis and then correct the output. Incorrect words are inside of tags '$<$bad$>$ $<$/bad$>$'. Please use this feedback in your correction.  If the hypothesis is already correct, do not make any changes. \\

\textcolor{teal}{English: Cookies are part of the HTTP header, so setcookie() must be called before any output is sent to the browser.}\\
\textcolor{blue}{Hypothesis: Cookies sind Teil des HTTP-Headers , deshalb muss setcookie() vor jedem \textbf{$<$bad$>$} Ausgabe-Output \textbf{$<$/bad$>$} an den Browser aufgerufen werden .}\\
\textcolor{purple}{German: Sie sind Bestandteil des HTTP-Headers, was bedeutet, dass die Funktion setcookie() aufgerufen werden muss, bevor irgendeine Ausgabe an den Browser erfolgt.}\\

\textcolor{teal}{English: session.use\_only\_cookies specifies whether the module will only use cookies to store the session id on the client side.}\\
\textcolor{blue}{Hypothesis: session .use\_only\_cookies bestimmt , ob das Modul nur \textbf{$<$bad$>$} mit \textbf{$<$/bad$>$} Cookies die Session-ID auf dem \textbf{$<$bad$>$} Client-Betriebssystem speichert \textbf{$<$/bad$>$} .}\\
\textcolor{purple}{German: session.use\_only\_cookies spezifiziert, ob das Modul nur Cookies verwendet, um die Session-ID clientseitig zu speichern.}\\

\textcolor{teal}{English: Note that SID is only defined if the client didn't send the right cookie.}\\
\textcolor{blue}{Hypothesis: \textbf{$<$bad$>$} Beachtet \textbf{$<$/bad$>$} , dass SID nur definiert \textbf{$<$bad$>$} ist \textbf{$<$/bad$>$} , wenn der Client nicht den richtigen Cookie gesendet hat .}\\
\textcolor{purple}{German: Beachten Sie, dass SID nur definiert ist, wenn vom Client nicht das richtige Cookie gesendet wurde.}\\

\textcolor{teal}{English: The server does not support the request type of the body.}\\
\textcolor{blue}{Hypothesis: Der Server unterstützt nicht \textbf{$<$bad$>$} die \textbf{$<$/bad$>$} Anforderungstyp \textbf{$<$bad$>$} der \textbf{$<$/bad$>$} Body .}\\
\textcolor{purple}{German: Der Server unterstützt den angeforderten Typ nicht.\%1: request type}\\

\textcolor{teal}{English: Must be in \& active session on local console}\\
\textcolor{blue}{Hypothesis: Muss in \& aktiver Sitzung auf \textbf{$<$bad$>$} dem \textbf{$<$/bad$>$} lokalen Konsole}\\
\textcolor{purple}{German: Nur in \& aktiver Sitzung auf lokaler Konsole}\\

\textcolor{teal}{English: Like other headers, cookies must be sent before any output from your script (this is a protocol restriction).}\\
\textcolor{blue}{Hypothesis: Wie andere \textbf{$<$bad$>$} Headern \textbf{$<$/bad$>$} müssen Cookies vor jedem \textbf{$<$bad$>$} Ausgabe-Output \textbf{$<$/bad$>$} ( dies ist eine Protokoll-Einschränkung ) gesendet werden .
}\\
\textcolor{purple}{German:}\\

\end{tcolorbox}

\caption{Example of 5-shot prompt for English-to-German Post-Editing with error markings (\textit{MRK}). Error markings inside by tags \textbf{$<$bad$>$ $<$/bad$>$}. Each demonstration example consists of a source segment in English (in green), a translation hypothesis in German (in blue), and a reference translation (in red).}
\label{fig:prompt-example-mrk}
\end{figure*}

\end{document}